\documentclass[11pt]{article}

\usepackage[final]{acl}
\usepackage[table]{xcolor}  
\usepackage{times}
\usepackage{latexsym}
\usepackage{booktabs}
\usepackage{amsmath}
\usepackage[most]{tcolorbox}
\usepackage{xcolor}
\usepackage{enumitem}
\usepackage{tabularx}
\usepackage{array}

\newcommand{\codechip}[1]{\texttt{\small #1}}

\usepackage[T1]{fontenc}

\usepackage[utf8]{inputenc}

\usepackage{microtype}

\usepackage{inconsolata}

\usepackage{graphicx}

\title{BioTool: A Comprehensive Tool-Calling Dataset for Enhancing Biomedical Capabilities of Large Language Models}

\author{
  Xin Gao\textsuperscript{1}\thanks{\ Equal contribution.} \quad
  Ruiyi Zhang\textsuperscript{1}\footnotemark[1] \quad
  Meixi Du\textsuperscript{1} \quad
  Peijia Qin\textsuperscript{1} \quad
  Pengtao Xie\textsuperscript{1, 2}\thanks{\ Corresponding authors.} \\
  \\\
  \textsuperscript{1}UC San Diego \quad
  \textsuperscript{2}MBZUAI \\
  \texttt{\{xig022, ruz048, p1xie\}@ucsd.edu}
}

\begin{document}
\maketitle
\begin{abstract}
Despite the success of large language models (LLMs) on general-purpose tasks, their performance in highly specialized domains such as biomedicine remains unsatisfactory. A key limitation is the inability of LLMs to effectively leverage biomedical tools, which clinical experts and biomedical researchers rely on extensively in daily workflows. While recent general-domain tool-calling datasets have substantially improved the capabilities of LLM agents, existing efforts in the biomedical domain largely rely on in-context learning and restrict models to a small set of tools.  
To address this gap, we introduce \textsc{BioTool}, a comprehensive biomedical tool-calling dataset designed for fine-tuning LLMs. \textsc{BioTool} comprises 34 frequently used tools collected from the NCBI, Ensembl, and UniProt databases, along with 7,040 high-quality, human-verified query–API call pairs spanning variation, genomics, proteomics, evolution, and general biology. Fine-tuning a 4-billion-parameter LLM on \textsc{BioTool} yields substantial improvements in biomedical tool-calling performance, outperforming cutting-edge commercial LLMs such as GPT-5.1. Furthermore, human expert evaluations demonstrate that integrating a \textsc{BioTool}-fine-tuned tool caller significantly improves downstream answer quality compared to the same LLM without tool usage, highlighting the effectiveness of \textsc{BioTool} in enhancing the biomedical capabilities of LLMs. The full dataset and evaluation code are available at \url{https://github.com/gxx27/BioTool}.
\end{abstract}

\section{Introduction}

\begin{figure*}[t]
    \centering
    \includegraphics[width=\textwidth]{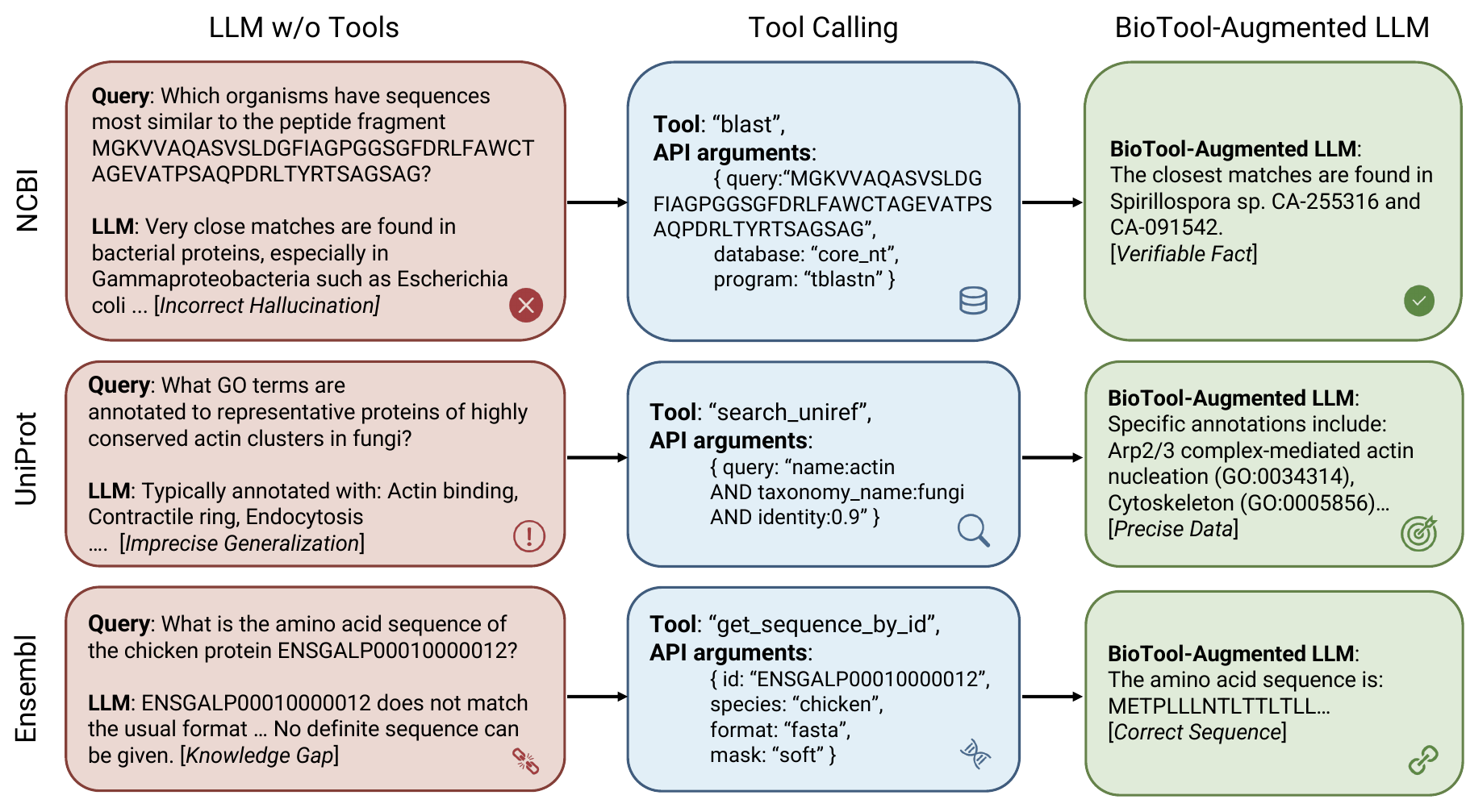}
    \caption{Comparison between answers generated by LLMs without tools and \textsc{BioTool}-augmented LLMs for biomedical queries. LLMs without tools often hallucinate or produce imprecise answers (left), whereas \textsc{BioTool}-augmented LLMs (right) generate API calls and retrieve critical information from biomedical databases, leading to higher-quality responses.}
    \label{fig:motivation}
\end{figure*}

The rapid advancement of large language models (LLMs) has revolutionized natural language processing, enabling unprecedented performance across a wide range of general-purpose tasks~\citep{Achiam2023GPT4TR, Bai2023QwenTR}. However, their capabilities in biomedical domains remain limited, which hinders their deployment in high-stakes, real-world biomedical applications~\citep{Chen2025Benchmarking, Li2025Benchmarking}. A key reason for this limitation is the insufficient ability of LLMs to effectively leverage specialized biomedical tools~\citep{jin2024genegpt}. Unlike commonsense questions that can often be answered directly, biomedical problems typically require even expert researchers to consult external tools and databases before drawing reliable conclusions~\citep{NCBI}. For instance, even for human biologists, the biological function of a raw nucleotide sequence cannot be reliably inferred without the aid of computational tools, such as BLAST or other sequence similarity–based methods~\citep{Altschul1990BLAST}. As shown in Figure~\ref{fig:motivation}, LLMs that lack access to or integration with such tools are therefore prone to hallucinations and imprecise generalizations, undermining their reliability for scientific discovery.

Given these challenges, early attempts have integrated biomedical and chemistry tools into LLMs via in-context learning~\citep{jin2024genegpt,bran2023chemcrow}. Although these approaches show improvements, they are constrained to a small set of available tools due to limited context length. Moreover, biomedical research tools often support diverse and complex usage scenarios that cannot be fully captured by a few lines of textual prompts, which hinders LLMs from fully realizing their potential in biomedical tool usage. Furthermore, they require models to map natural-language questions to highly specialized schemas, identifiers, and parameter conventions to reliably retrieve biologically relevant evidence. Inspired by the success of instruction-tuning–based tool-calling datasets in the general NLP domain~\citep{liu2024apigen,patil2024gorilla}, we address this gap by curating a comprehensive biomedical tool-calling dataset, \textsc{BioTool}.

\textsc{BioTool} is an instruction fine-tuning–style biomedical tool-calling dataset consisting of 7,040 high-quality, human-verified query–API call pairs. It includes 34 frequently used tools from the NCBI~\citep{NCBI}, Ensembl~\citep{Hubbard2002Ensembl}, and UniProt~\citep{UniProt2017Knowledgebase} databases, spanning multiple subdomains such as variation, genomics, proteomics, evolution, and general biology.
To construct the dataset, we first manually select 34 tools from NCBI, Ensembl, and UniProt that are widely used in biomedical research. We then collect official documentation for these tools from their respective websites and use them to generate diverse combinations of API parameters with the assistance of LLMs. The synthesized API calls are executed and filtered to remove cases with unavailable or uninformative responses, resulting in 3,829 unique API calls. 
Next, we prompt cutting-edge reasoning models~\citep{openai2025o3o4mini} with these API calls and their corresponding responses to generate potential user queries. These queries are subsequently evaluated by an LLM-based judge to assess whether the API responses meaningfully support answering the queries, followed by a final round of human expert review focusing on biological relevance and correctness. This process yields 7,040 high-quality query–API call pairs, which is the final \textsc{BioTool} dataset.

We evaluate the quality and effectiveness of \textsc{BioTool} through two sets of experiments. First, we fine-tune several open-source LLMs with 4B to 8B parameters on the \textsc{BioTool} training split and compare them with cutting-edge commercial LLMs, including GPT-5.1, Gemini-3 Pro, and Claude-4.5-Sonnet, using in-context learning. Results on the test split show that smaller LLMs fine-tuned with \textsc{BioTool} significantly outperform commercial LLMs with hundreds of times more parameters in terms of tool-calling quality. For example, a \textsc{BioTool}-fine-tuned 4B Qwen-3 model outperforms the best-performing Claude-4.5-Sonnet by 15.0\% in overall API-calling quality.
Second, we conduct human evaluations to assess whether \textsc{BioTool}-enhanced LLMs produce higher-quality answers from the perspective of biomedical researchers. On 1,048 test queries, a GPT-5.1 model augmented with oracle \textsc{BioTool} API calls achieves 88.4\% higher normalized answer quality compared to the same model without tool usage, demonstrating the intrinsic quality of the \textsc{BioTool} dataset. Moreover, a GPT-5.1 model augmented with a \textsc{BioTool}-fine-tuned API caller achieves 69\% higher normalized answer quality compared to the raw GPT-5.1 model, highlighting the effectiveness of \textsc{BioTool} in training tool-using LLMs and enhancing their biomedical capabilities.

\section{Related Works}

Early general-purpose tool-calling models, such as Toolformer~\cite{schick2023toolformerlanguagemodelsteach} and Gorilla~\cite{patil2024gorilla}, established that LLMs can be trained to invoke external APIs, thereby grounding responses in retrieved data to mitigate hallucinations. Subsequent frameworks like ToolBench~\cite{qin2023toolllm} and APIGen~\cite{liu2024apigen} advanced this capability by introducing scalable pipelines for generating synthetic instruction-tuning data. Despite these advancements, generalist models often struggle with specialized scientific domains like biomedicine because they rely on broad datasets that include only a negligible fraction of corresponding tools and frequently fail to adhere to the rigorous schema constraints of scientific databases. To address these limitations, domain-specific agents have emerged. GeneGPT~\cite{jin2024genegpt} pioneered this shift by utilizing in-context learning~\cite{wei2023chainofthoughtpromptingelicitsreasoning} to enable access to NCBI Web APIs. Similarly, systems such as SciAgent~\cite{li2025sciagentunifiedmultiagentgeneralistic} and ChemCrow~\cite{bran2023chemcrow} have successfully integrated tool-augmented agents for complex reasoning in scientific and chemical research. While more recent entries like Biomni~\cite{huang2025biomni} have introduced general-purpose agents for biomedical tasks, they primarily focus on a restricted subset of tools. Consequently, they lack the comprehensive, full-list interface to primary authoritative biomedical databases.

\begin{figure*}
    \centering
    \includegraphics[width=1.0\textwidth]{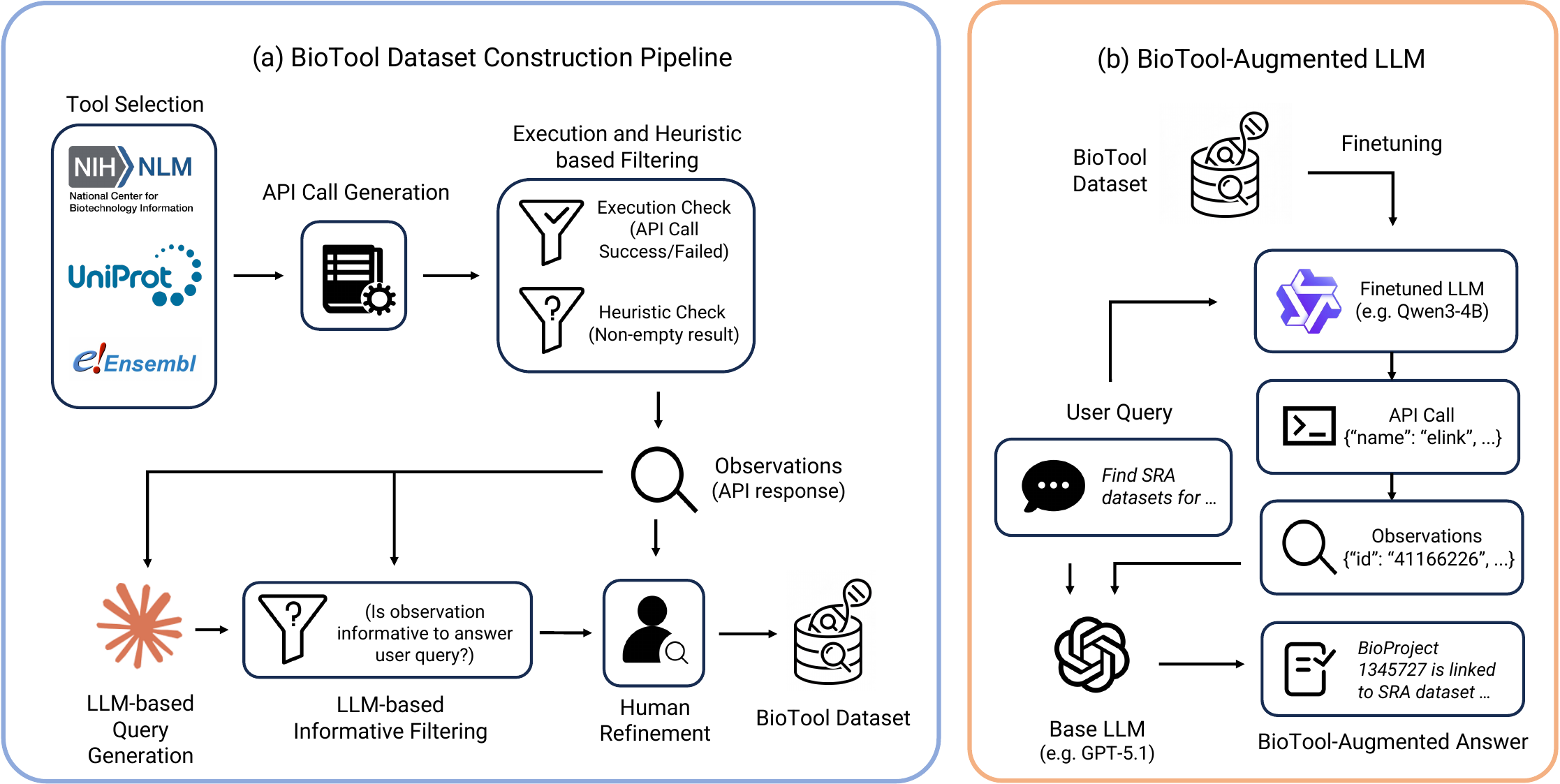}
    \caption{The systematic workflow of \textsc{BioTool} spans from automated dataset construction to downstream application. Panel (a) illustrates the multi-stage construction pipeline, which includes initial tool selection from primary databases, automated API call generation, and a rigorous filtering process involving execution checks, heuristic validation, and LLM-based informativeness assessment. Panel (b) depicts the inference-time application, where specialized API-calling models fine-tuned on \textsc{BioTool} enable base LLMs to retrieve grounded observations and generate verifiable biological answers.}
    \label{fig:procedure}
\end{figure*}

\section{The BioTool Dataset}

This section details the development and composition of \textsc{BioTool}. 
We first present an example data entry from \textsc{BioTool} to illustrate the structure of a query–API call pair. Each entry includes a \textit{user query} field, which contains a realistic clinical or biomedical question expressed in free-form text. The \textit{tool information} field provides descriptions of the tools required to answer the query, while the \textit{API arguments} specify the input parameters for the corresponding API endpoint. Executing the API endpoint with these arguments returns an \textit{observations}, which contains information used to augment the LLM’s response. We note that the observation is fully determined by the API endpoint and its arguments; it is included in the dataset for completeness and user convenience. 

Next, we describe the sequential construction pipeline used to generate and verify biomedical tool calling pairs in Section \ref{sec:data_construction}, illustrated in Figure~\ref{fig:procedure}. We then provide a quantitative analysis of the resulting dataset, highlighting its functional utility and biological diversity in Section \ref{sec:data_statistics}.

\newtcolorbox{datacollection}[1]{%
  enhanced,
  breakable,
  colback=gray!0,
  colframe=gray!90,
  colbacktitle=gray!10,
  coltitle=black,
  title={#1},
  fonttitle=\sffamily,
  boxrule=1.5pt,
  leftrule=1.5pt,
  rightrule=1.5pt,
  toprule=1.5pt,
  bottomrule=1.5pt,
  left=3.5mm,
  right=3.5mm,
  top=3.5mm,
  bottom=3.5mm,
  width=\columnwidth,
  fontupper=\sffamily\small,
  before upper={\setlength{\parskip}{2pt}},
  before skip=8pt,
  after skip=8pt,
}

\begin{datacollection}{\small Example of BioTool Data Entry}

\noindent\textbf{User Query}\par
\codechip{Could you provide concise definitions for the major severe immunodeficiency disorders?}

\vspace{0.6em}
\noindent\textbf{Tool Information}\par
\small
\setlength{\tabcolsep}{4pt}
\renewcommand{\arraystretch}{1.15}
\begin{tabularx}{\linewidth}{@{}>{}l @{\hspace{0.6em}} X@{}}
\codechip{Database:} & \codechip{``UniProt''} \\
\codechip{Tool:} & \codechip{``human\_diseases''} \\
\codechip{API Endpoint:} & \codechip{``search\_human\_diseases''} \\
\end{tabularx}

\vspace{0.6em}
\noindent\textbf{API Arguments}\par
\small
\begin{tabularx}{\linewidth}{@{}>{}l @{\hspace{0.6em}} X@{}}
\codechip{query:}  & \codechip{``name: immunodeficiency AND name: severe''} \\
\codechip{fields:} & \codechip{``definition''} \\
\codechip{sort:}   & \codechip{``id asc''} \\
\end{tabularx}

\vspace{0.6em}
\noindent\textbf{Observations}\par
\small
\setlength{\tabcolsep}{4pt}
\renewcommand{\arraystretch}{1.15}
\begin{tabularx}{\linewidth}{@{}X@{}}
\codechip{(id: ``DI-00171'', definition: ``An autosomal recessive immunologic disorder characterized by the loss of expression of MHC class II antigens on antigen-presenting cells...''),}\\
\codechip{(id: ``DI-00305'', definition: ``A form of chronic granulomatous disease...''),}\\
\codechip{...}\\
\end{tabularx}
\end{datacollection}

\begin{figure*}[t]
    \centering
    \includegraphics[width=\textwidth]{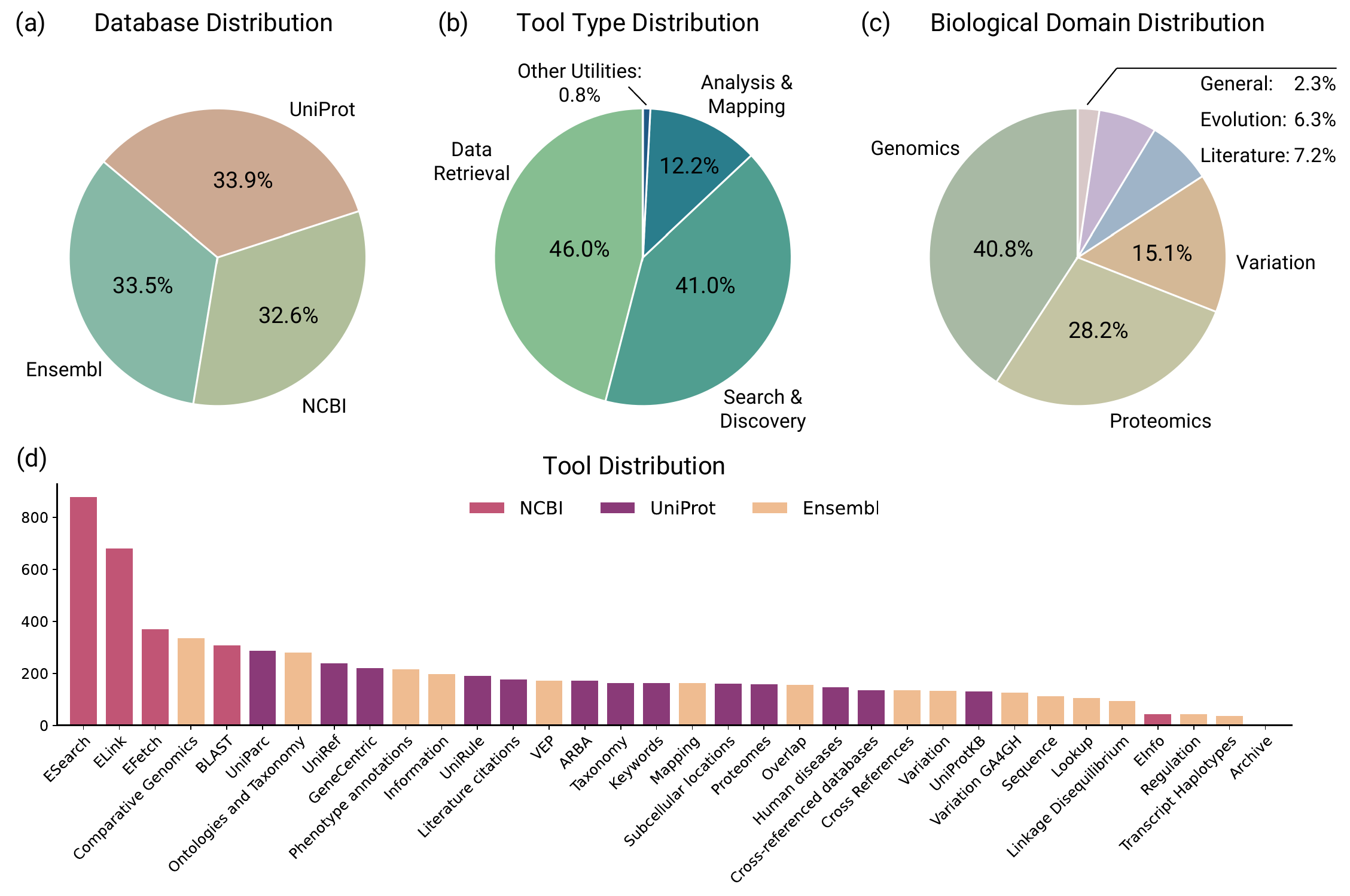}
    \caption{Distribution analysis of the 7,040 samples within \textsc{BioTool} across four dimensions. Panel (a) shows the distribution across source databases. Panel (b) illustrates the distribution of samples by tool type. Panel (c) presents the distribution across various biological domains. Panel (d) delineates the distribution of user queries across the 34 distinct biological tools.}
    \label{fig:statistics}
\end{figure*}

\subsection{Dataset Construction Pipeline}\label{sec:data_construction}

\paragraph{Tool Selection}
We select three major online API providers: the National Center for Biotechnology Information (NCBI), UniProt, and Ensembl as the tool source for \textsc{BioTool}, motivated by their roles as the authoritative repositories within the global biomedical research infrastructure~\cite{sayers2010general, 10.1093/nar/gkaf394, 10.1093/bioinformatics/btu613}. These three platforms are widely considered the definitive standard because they offer expansive and highly interoperable data spanning the entire central dogma of biology, encompassing the full spectrum from raw genomic sequences to functional protein annotations.

Across the three databases, we comprehensively review their websites and manually select tools that are critical for answering biomedical and clinical questions. During this process, we exclude tools with limited biomedical relevance (e.g., APIs that only return service or versioning information) as well as deprecated or unstable tools. As a result, we curate a diverse set of 34 tools comprising 124 API endpoints, each of which is frequently used in biomedical research workflows. The complete list of selected tools is provided in Appendix~\ref{sec:appendix-tool}. 
In addition, we collect the official documentation for each API endpoint from the corresponding website. These documents specify API usage, input arguments, constraints, and example calls, and serve as essential resources for subsequent stages of API call synthesis and user query generation.

\paragraph{API Call Synthesis and Verification}
Based on the curated tool set and associated documentation, we manually select critical API arguments corresponding to biologically meaningful identifiers for each API endpoint. These arguments, such as taxon IDs, gene symbols, and UniProt accession numbers, ensure that the synthesized API calls are biologically diverse and scientifically plausible. Given the selected arguments, we follow prior work~\citep{liu2024apigen} to randomly sample a large set of candidate API calls. These candidates are then executed to filter out cases that result in client errors, timeouts, or empty responses.
To further improve data quality, we design a novel heuristic-based filtering strategy to remove API calls that are overly similar to existing ones, as well as those whose returned observations lack biological significance. Details of this heuristic filter are provided in Appendix~\ref{sec:appendix_heuristic}. After this verification process, we obtain a collection of 6,391 unique API calls.

\paragraph{User Query Generation}
Given the synthesized API calls, we leverage cutting-edge LLMs to generate corresponding user queries, following a self-instruct–style paradigm established in prior work~\citep{Wang2022SelfInstructAL,patil2024gorilla,liu2024apigen}. Specifically, LLMs are prompted with an API call, its documentation, and its corresponding observation, together with a small set of human-crafted in-context query–API call pairs, to generate realistic user queries.

To further improve the quality and biological relevance of \textsc{BioTool}, we introduce two novel adaptations to ensure both the \textit{necessity} and \textit{sufficiency} of the API observations. First, to enforce \textit{necessity}, we apply Chain-of-Thought (CoT) prompting~\citep{wei2023chainofthoughtpromptingelicitsreasoning} using a strong reasoning model (OpenAI o3~\citep{openai2025o3o4mini}) when generating user queries. The model is first prompted to summarize the technical details of the API observation into a natural-language description, which is then used to generate the final user query. This procedure ensures that the observation is required to answer the query, while keeping the query realistic and avoiding explicit references to specific tools or API calls. The detailed system and user prompts for this process are provided in Appendix~\ref{sec:appendix_user_queries_prompt}.
Second, to ensure \textit{sufficiency}, we employ another cutting-edge LLM (Claude Haiku 4.5~\citep{anthropic2025haiku45}) to perform informativeness-based filtering, inspired by the LLM-as-a-judge framework~\citep{zheng2023judgingllmasajudgemtbenchchatbot}. The model is prompted to follow a structured rubric and classify a query–API call pair as informative if the observation contains at least one relevant fact or a partial summary that supports the user’s intent. Pairs in which the observation is unrelated to the query or too vague to support a concrete response are discarded. The specific judge prompts are provided in Appendix~\ref{sec:appendix_informative_check_prompt}.

\paragraph{Human Refinement}
The final stage involves a comprehensive manual review conducted by human evaluators with at least a college-level background in bioinformatics. The evaluators first identify and remove low-quality queries. For the remaining samples, they refine pedantic or unnatural phrasing and ensure the accuracy of biological terminology and nomenclature. After this round of filtering and correction, the final \textsc{BioTool} dataset comprises 7,040 high-quality samples.

This instruction fine-tuning–style dataset is primarily used to train open-source LLMs as API-calling models, following training paradigms established in general-domain tool-calling datasets~\citep{patil2024gorilla,liu2024apigen}. A \textsc{BioTool}-trained LLM can assist state-of-the-art LLMs in generating grounded and scientifically accurate responses, as illustrated in the right panel of Figure~\ref{fig:procedure}.

\subsection{Data Statistics}\label{sec:data_statistics}

The \textsc{BioTool} dataset is derived from 34 distinct biological tools and 124 unique API endpoints, encompassing a wide array of scientific content categorized across several key dimensions. As shown in Figure~\ref{fig:statistics}(a), the distribution of tools across databases is well balanced, with comparable proportions from NCBI, UniProt, and Ensembl. Figure~\ref{fig:statistics}(b) illustrates the diversity of tool types included in \textsc{BioTool}, ranging from data retrieval (e.g., nucleotide identifiers fetching) and search and discovery (e.g., phenotype-based gene discovery) to biological analysis and mapping (e.g., cross-referencing SNP identifiers). Figure~\ref{fig:statistics}(c) highlights the dataset’s broad scientific scope, covering domains such as genomics (e.g., gene tree querying), proteomics (e.g., protein sequence alignment), variation analysis (e.g., linkage disequilibrium analysis), and evolutionary biology (e.g., species-level taxonomy identification). Finally, Figure~\ref{fig:statistics}(d) shows that \textsc{BioTool} includes both frequently accessed general-purpose tools and a long tail of specialized tools, all of which are essential for complex scientific discovery across the central dogma.

\section{Experimental Results}

To evaluate the effectiveness of \textsc{BioTool}, we first compare the API-calling capabilities of small open-source LLMs fine-tuned on \textsc{BioTool} against their vanilla counterparts and cutting-edge proprietary LLMs using in-context learning. We then conduct human expert evaluations to compare the answer quality of baseline LLMs with that of \textsc{BioTool}-augmented LLMs.

\subsection{Experimental Setup}

\paragraph{BioTool score} We define a BioTool performance score to automatically evaluate the capability of an LLM as an API caller on the \textsc{BioTool} dataset, especially the alignment of retrieved information with the user's intent. Specifically, assume we have the test set $D=\{(q_1,o_1),...(q_n, o_n)\}$, where $q_i$ is the $i^{\text{th}}$ user query and $o_i$ is the observation obtained from ground-truth API calling in the dataset. The BioTool score on this test set $\mathrm{S}(D)$ for a LLM API caller $f$ is then defined as follows:

\begin{equation}
    \mathrm{S}(D)=\sum_{i=1}^n\mathrm{Sim}\bigl(f(q_i), o_i\bigr)
\end{equation}
where $\mathrm{Sim}(\hat{o}, o)$ computes the semantic embedding similarity of two text strings: the ground truth observation $o$ and the corresponding observation $\hat{o}$ from LLM API caller prediction. In practice, we use a MedCPT model~\citep{jin2023medcpt} to get a sentence embedding for an observation. API calls may fail due to incorrect model generation, yielding an empty string $\hat{o} = \varepsilon$. In this case, we set $\mathrm{Sim}(\varepsilon, o) = 0$. Intuitively, this score determines model performance by measuring whether the retrieved biological facts remain semantically similar to the required information, even when the technical implementation of the call differs from the reference.

\paragraph{Additional Metrics}
Based on the BioTool score, we define two additional metrics to further characterize model performance. Similar metrics have been widely adopted in existing API-calling benchmarks~\citep{patil2025the}. Firstly, we define API calling success rate $\mathrm{AS}$ as follows:

\begin{equation}
\mathrm{AS}(D)
= \frac{1}{n} \sum_{i=1}^{n}
\mathbf{1}\!\left[\mathrm{Sim}\bigl(f(q_i), o_i\bigr) > 0\right]
\end{equation}
where $\mathbf{1}\left[\cdot\right]$ is the indicator function. A zero similarity indicates API calling failure due to incorrect formatting, invalid API names, or improper parameter values. Conceptually, this metric focuses on the model's capability to generate API calls that execute correctly and return a valid response containing data. Secondly, we define a exact match score $\mathrm{EM}$ as follows:

\begin{equation}
\mathrm{EM}(D)
= \frac{1}{n} \sum_{i=1}^{n}
\mathbf{1}\!\left[\mathrm{Sim}\bigl(f(q_i), o_i\bigr) = 1\right]
\end{equation}
which measures the proportion of predictions whose resulting observations exactly match the ground-truth reference observation, requiring the model to correctly identify the API endpoint and provide all required parameters with values that exactly match the reference. 

\paragraph{Models} In this study, we use four cutting-edge proprietary models, including GPT-5.1, GPT-5.1-Codex, Gemini 3 Pro, and Claude 4.5 Sonnet~\cite{openai2025gpt51addendum, openai2025gpt51codex, google2025gemini, anthropic2025claude} under an in-context learning scheme. We use four open-source models, which are Llama3.1-8B-Instruct, Qwen3-8B, Qwen2.5-7B-Instruct, and Qwen3-4B-Instruct~\cite{grattafiori2024llama3herdmodels, yang2025qwen3technicalreport, qwen2025qwen25technicalreport}, for both in-context learning and BioTool-based fine-tuning. We report the average performance across three independent runs.

\subsection{Results on Tool Calling Capability}\label{sec:experimental_results}

In this section, we first fine-tune small open-source models on the training split of the \textsc{BioTool} dataset, which is randomly split under a four-to-one ratio. We use the cutting-edge proprietary model and base open-source models as baselines, and the evaluation for all models was conducted equally on the held-out test set consisting of 1,408 samples in terms of BioTool score. As shown in Table~\ref{tab:results_cs}, there is a clear performance advantage for \textsc{BioTool}-fine-tuned models over much larger LLMs under in-context learning. The fine-tuned 4B model achieved the highest overall BioTool score, representing a 15.0\% improvement over the strongest proprietary model, Claude 4.5 Sonnet, and 68.9\% higher performance than GPT-5.1. This gap suggests that the general-purpose pre-training of frontier LLMs together with in-context learning is insufficient to navigate the specialized technical constraints and precise parameter mappings of biological repositories. Instead, the high-density training signals within the \textsc{BioTool} dataset allow significantly smaller models to acquire the necessary domain expertise that remains elusive to even the largest proprietary models.

\begin{table}[t]
\centering
\setlength{\tabcolsep}{2pt}

\begin{tabular}{lcccc}
\toprule
{\textbf{Model}} & 
{\small\textbf{NCBI}} & 
{\small\textbf{UniProt}} & 
{\small\textbf{Ensembl}} & 
{\small\textbf{Overall}} \\
\midrule

\rowcolor{gray!15}
\multicolumn{5}{c}{Proprietary Models}\vspace{0.1cm} \\
GPT-5.1           & 15.5 & 78.7 & 72.8 & 55.4 \\
GPT-5.1-Codex     & 14.8 & 80.5 & 74.5 & 56.3 \\
Gemini 3 Pro      & 72.5 & 87.8 & 77.3 & 79.2 \\
Claude 4.5 Sonnet & 79.8 & 87.3 & 77.0 & 81.4 \\

\midrule
\rowcolor{gray!15}
\multicolumn{5}{c}{Base Open-Source Models}\vspace{0.1cm} \\
Llama3.1-8B-Ins & 28.4 & 77.5 & 58.8 & 54.8 \\
Qwen3-8B & 26.2 & 73.1 & 63.5 & 54.1 \\
Qwen2.5-7B-Ins & 48.0 & 79.0 & 65.3 & 64.0 \\
Qwen3-4B-Ins & 53.4 & 71.9 & 64.0 & 63.1 \\

\midrule
\rowcolor{gray!15}
\multicolumn{5}{c}{BioTool-fine-tuned Models}\vspace{0.1cm} \\
Llama3.1-8B-Ins & 91.1 & 88.9 & 93.7 & 91.2 \\
Qwen3-8B & 89.5 & 84.8 & 83.9 & 86.1 \\
Qwen2.5-7B-Ins & 91.8 & 87.6 & 88.0 & 89.2 \\
Qwen3-4B-Ins & \textbf{93.7} & \textbf{91.0} & \textbf{96.3} & \textbf{93.6} \\





\bottomrule
\end{tabular}

\caption{Comparative evaluation of models on the \textsc{BioTool} dataset, measured by the BioTool score (higher is better). Scores are reported for each constituent database (NCBI, UniProt, Ensembl) and overall. Model names with the suffix \emph{Ins} denote instruction-tuned variants. Bold values indicate the best performance in each column.}
\label{tab:results_cs}
\end{table}

\subsection{Human Evaluation of Answer Quality}\label{sec:results_human_evaluation}

\begin{figure}[t]
    \centering
    \includegraphics[width=\linewidth]{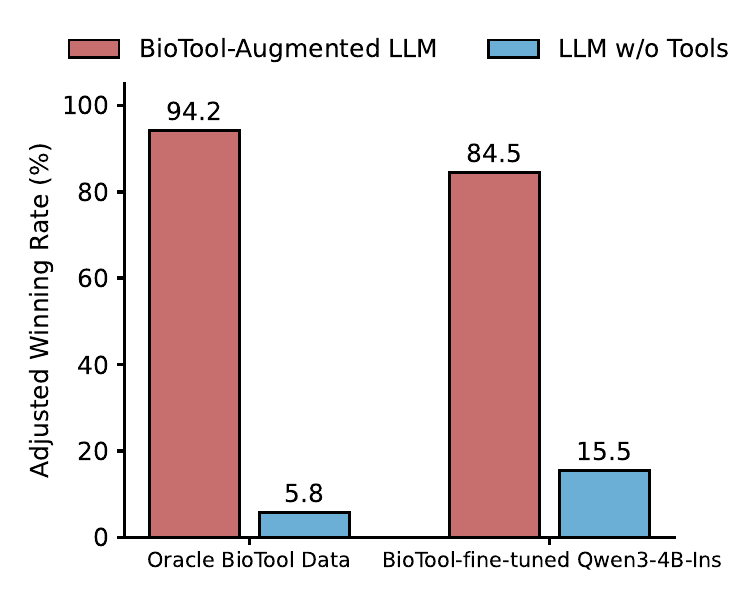}
    \caption{Human evaluation results comparing answer quality between \textsc{BioTool}-augmented LLMs and LLMs without tool usage, using GPT-5.1 as the base model. The augmented settings include GPT-5.1 with oracle \textsc{BioTool} data (left) and GPT-5.1 with a \textsc{BioTool}-fine-tuned Qwen3-4B-Instruct tool caller (right).}
    \label{fig:human_eval}
\end{figure}

The ultimate criterion for assessing the usefulness of a tool-calling dataset is its ability to improve the quality of LLM-generated answers. To evaluate this, we use GPT-5.1 as the base model and compare its performance under three settings: (1) no tool augmentation, (2) augmentation with ground-truth \textsc{BioTool} API calls, and (3) augmentation with a BioTool-fine-tuned Qwen3-4B-Instruct tool-calling model. We evaluate these three settings across all test queries using side-by-side human judgments by two annotators with college-level bioinformatics backgrounds. Annotators compare settings (1) vs. (2) and (1) vs. (3), selecting the better answer based on informativeness and task fulfillment, while rejecting vague or scientifically incorrect responses. The normalized win rates for the two comparisons are shown in Figure~\ref{fig:human_eval}. The reported win rates are the average of the two annotators’ individual results. Raw preference results and normalization procedures are detailed in Appendix~\ref{sec:appendix-human-eval}.

We observe that tool augmentation substantially improves the quality of biomedical answers, demonstrating that grounding LLMs in verifiable data from NCBI, Ensembl, and UniProt effectively mitigates domain-specific hallucinations and imprecise generalizations. The oracle configuration achieves a 94.2\% win rate over the base model, highlighting the high quality of the \textsc{BioTool} dataset. Similarly, the \textsc{BioTool}-fine-tuned Qwen3-4B-Instruct model attains an 84.5\% win rate, indicating that a small, fine-tuned model can improve the correctness and helpfulness of large commercial LLMs as judged by human evaluators, further demonstrating the practical utility of \textsc{BioTool}.

\begin{table*}[t]
\centering
\normalsize
\setlength{\tabcolsep}{10pt}

\begin{tabular}{lcccccccc}
\toprule
\textbf{Model} &
\multicolumn{2}{c}{\textbf{NCBI}} &
\multicolumn{2}{c}{\textbf{UniProt}} &
\multicolumn{2}{c}{\textbf{Ensembl}} &
\multicolumn{2}{c}{\textbf{Overall}} \\
& $\mathrm{EM}$ & $\mathrm{AS}$ & $\mathrm{EM}$ & $\mathrm{AS}$ & $\mathrm{EM}$ & $\mathrm{AS}$ & $\mathrm{EM}$ & $\mathrm{AS}$ \\
\midrule

\rowcolor{gray!15}
\multicolumn{9}{c}{{Proprietary Models}}\vspace{0.1cm} \\

GPT-5.1 & 0.0 & 16.6 & 1.7 & 97.9 & 8.9 & 79.4 & 3.5 & 64.4 \\
GPT-5.1-Codex & 0.0 & 16.2 & 1.7 & 98.3 & 6.3 & 80.5 & 2.6 & 64.7 \\
Gemini 3 Pro & 3.2 & 76.7 & 2.1 & 99.8 & 16.3 & 82.0 & 7.1 & 86.2 \\
Claude 4.5 Sonnet & 3.6 & 85.5 & 1.1 & \textbf{100.0} & 15.0 & 82.4 & 6.5 & 89.4 \\

\midrule
\rowcolor{gray!15}
\multicolumn{9}{c}{Base Open-Source Models}\vspace{0.1cm} \\

Llama3.1-8B-Ins & 0.0 & 33.2 & 1.0 & 96.7 & 5.5 & 68.9 & 2.1 & 66.1 \\
Qwen3-8B & 0.5 & 28.9 & 1.2 & 87.8 & 10.0 & 70.1 & 3.8 & 62.1 \\
Qwen2.5-7B-Ins & 0.0 & 54.6 & 1.6 & 97.6 & 6.3 & 74.1 & 2.6 & 75.4 \\
Qwen3-4B-Ins & 1.3 & 58.5 & 1.7 & 88.8 & 7.8 & 72.8 & 3.6 & 73.3 \\

\midrule
\rowcolor{gray!15}
\multicolumn{9}{c}{BioTool-fine-tuned Models}\vspace{0.1cm} \\

Llama3.1-8B-Ins & 43.1 & 93.3 & 7.4 & 99.9 & 43.1 & 96.6 & 31.2 & 96.6 \\
Qwen3-8B & 40.8 & 92.3 & 4.5 & 99.8 & 35.6 & 88.8 & 26.9 & 93.7 \\
Qwen2.5-7B-Ins & 44.6 & 93.8 & 5.4 & 99.9 & 42.4 & 91.5 & 30.8 & 95.1 \\
Qwen3-4B-Ins & \textbf{66.7} & \textbf{94.8} & \textbf{9.3} & \textbf{100.0} & \textbf{51.1} & \textbf{98.8} & \textbf{42.4} & \textbf{97.8} \\

\bottomrule
\end{tabular}

\caption{Comparative evaluation of EM and AS metrics (higher is better). Model names with the suffix \emph{Ins} denote instruction-tuned variants. Bold values indicate the best performance in each column.}
\label{tab:results_em_as}
\end{table*}

\subsection{Additional Results}
We report results for the additional metrics under the same experimental settings in Table~\ref{tab:results_cs} to provide further insights into model behavior and dataset characteristics. As shown in Table~\ref{tab:results_em_as}, there is a clear divergence between Exact Match (EM) and API Success (AS), particularly for proprietary models. Although models such as Claude 4.5 Sonnet and Gemini 3 Pro achieve high AS scores, their EM remains extremely low, indicating difficulty in producing parameterizations that exactly match reference specifications. In contrast, the \textsc{BioTool}-fine-tuned Qwen3-4B-Instruct achieves an EM nearly six times that of the best proprietary model, highlighting the necessity of fine-tuning for learning the precise syntax of biological APIs. The EM–AS gap also reflects the varying complexity of biological repositories. On the NCBI subset, proprietary models such as GPT-5.1 fail to achieve any exact matches and frequently encounter execution errors, likely due to strict identifier formats and nested parameters. Fine-tuned models, however, maintain high execution success, demonstrating that \textsc{BioTool} trains functionally robust models that produce valid and biologically meaningful API calls even without exact string matches.

\begin{table}[t]
\centering
\setlength{\tabcolsep}{4pt}
\begin{tabular}{lccc}
\toprule
\textbf{Model} & \textbf{Missing} & \textbf{Extra} & \textbf{Wrong} \\
\midrule
\rowcolor{gray!15}
\multicolumn{4}{c}{Proprietary Models}\vspace{0.05cm}\\
GPT-5.1           &  5.0 & 34.9 & 28.9 \\
GPT-5.1-Codex     &  5.2 & 34.6 & 30.0 \\
Gemini 3 Pro      &  5.8 &  8.2 & 10.9 \\
Claude 4.5        &  5.8 &  4.0 &  8.2 \\
\midrule
\rowcolor{gray!15}
\multicolumn{4}{c}{Base Open-Source Models}\vspace{0.05cm}\\
Llama3.1-8B-Ins   & 14.4 & 27.6 & 24.6 \\
Qwen3-8B          & 12.2 & 17.0 & 14.3 \\
Qwen2.5-7B-Ins    & 12.4 & 17.0 & 15.3 \\
Qwen3-4B-Ins      &  6.1 &  6.4 &  8.2 \\
\midrule
\rowcolor{gray!15}
\multicolumn{4}{c}{BioTool-fine-tuned Models}\vspace{0.05cm}\\
Llama3.1-8B-Ins   &  1.1 &  1.1 &  1.8 \\
Qwen3-8B          &  0.9 &  1.1 &  2.0 \\
Qwen2.5-7B-Ins    &  1.8 &  2.1 &  2.6 \\
Qwen3-4B-Ins      &  \textbf{0.6} &  \textbf{0.6} &  \textbf{1.6} \\
\bottomrule
\end{tabular}%
\caption{Distribution of parameter-level error categories among API failures. \textbf{Missing}, \textbf{Extra}, and \textbf{Wrong} denote the proportion of the total test set attributable to failures involving missing, extra, and wrong-value parameters, respectively. Bold values indicate the lowest error rate in each column.}
\label{tab:error_analysis}
\end{table}

\subsection{Error Analysis}

To gain deeper insight into model failure modes, we conduct a systematic error analysis over all API call failures on the whole test set. We categorize parameter-level mistakes into three mutually non-exclusive types. \textit{Missing parameters} refers to cases where the predicted call omits one or more arguments present in the ground-truth reference, thereby altering the biological scope of the retrieved result; for instance, the omission of the \texttt{species} argument from an Ensembl comparative genomics call, which causes the endpoint to return homology data for an unintended reference organism. \textit{Extra parameters} refers to cases where the predicted call includes arguments absent from the reference, potentially redirecting the query's intent; like injecting a \texttt{canonical} flag into a VEP annotation call, which restricts consequence reporting to canonical transcripts only and suppresses annotations for non-canonical isoforms that may be biologically or clinically relevant. \textit{Wrong parameter values} refers to cases where the argument name is correct, but the assigned value is semantically incorrect; for example, specifying \texttt{"blastp"} in place of \texttt{"tblastn"} as the BLAST program, which conflates protein-against-protein and protein-against-translated-nucleotide search modes and yields entirely incompatible results. The distribution of these error types across all evaluated models is reported in Table~\ref{tab:error_analysis}.

As shown in the results, proprietary models and non-fine-tuned open-source models exhibit pervasive failures in semantic parameter mapping, including incorrect database or program selections in NCBI Entrez and BLAST calls, misspecified biological identifiers such as taxonomic names, and erroneous traversal targets in Entrez link operations, all of which cause the retrieved data to be biologically misaligned with the user's intent, regardless of whether the call itself executes successfully. \textsc{BioTool} fine-tuning substantially mitigates these semantic mapping failures, yielding dramatically reduced API failure rates across all trained model variants and confirming that correct biomedical API invocation requires domain-specific schema grounding that neither general-purpose pretraining nor in-context learning can reliably provide.



\subsection{Case Study}

To illustrate how \textsc{BioTool} enhances LLMs’ biomedical capabilities, we present an example comparing the API-calling behavior and tool-augmented answers of a \textsc{BioTool}-fine-tuned Qwen3-4B-Instruct model and Claude 4.5 Sonnet on a complex genomic linkage query. The ICL–based Claude model exhibits low precision in handling domain-specific tool usage rules, directly passing literal natural-language terms such as ``goat'' and ``NextGen'' from the user query into the API arguments. In contrast, the \textsc{BioTool}-fine-tuned model correctly maps these concepts to the required API parameters. This example demonstrates that \textsc{BioTool} provides high-quality supervision for learning the implicit rules of biomedical tool usage, enabling reliable natural-language-to-tool mapping that state-of-the-art ICL-based LLMs often fail to achieve.

\begin{datacollection}{\small Comparison of API Call and Response to an Example User Query}

\noindent\textbf{User Query}\par
\codechip{Within 29:707234-757234, which variant pairs show strong linkage in the NextGen goat population?}

\vspace{0.6em}
\noindent\textbf{BioTool-fine-tuned Qwen3-4B API Call}\par
\codechip{get\_ld\_region(species=``capra\_hircus'', region=``29:707234-757234'', population\_name=``NextGen:All'', d\_prime=0.8)}

\vspace{0.6em}
\noindent\textbf{BioTool-fine-tuned Qwen3-4B Response}\par
\codechip{[ \{"variation1": ``rs661133063'', ``variation2'': ``rs668584442'', ``d\_prime'': 1.0 \}, ... ]}

\vspace{0.6em}
\noindent\textbf{Claude API Call}\par
\codechip{get\_ld\_region(species=``goat'', region=``29:707234-757234'', population\_name=``NextGen'')}

\vspace{0.6em}
\noindent\textbf{Claude Response}\par
\codechip{\{ ``error'': ``Can not find internal name for species 'goat' '' \}}

\vspace{0.6em}
\noindent\textbf{Oracle API Call}\par
\codechip{get\_ld\_region(species=``capra\_hircus'', region=``29:707234-757234'', population\_name=``NextGen:All'', r2=0.5)}

\end{datacollection}

\section{Conclusion}

In this work, we introduce \textsc{BioTool}, a comprehensive biomedical tool-calling dataset comprising 7,040 human-verified query-API call pairs spanning 124 biomedical tools. Fine-tuning a 4-billion-parameter LLM on \textsc{BioTool} leads to substantial improvements in API-calling performance, surpassing cutting-edge proprietary LLMs. Furthermore, human evaluations confirm that \textsc{BioTool}-augmented LLMs generate more helpful, informative, and scientifically accurate answers compared to the same base models without tool usage, shedding light on the development of reliable biomedical agents in the future.

\section*{Limitations}

Despite the performance gains observed with \textsc{BioTool}, several limitations remain. Our current framework focuses exclusively on one-hop tool calling responses. This ignores more complex biological problems that cannot be solved with a single API interaction and instead require multi-hop search results or iterative reasoning across multiple tools. Furthermore, we did not fine-tune an independent, specialized biomedical agent. This architectural choice was necessitated by the extreme context length of raw biological observations, which frequently exceed our resource limitations even after post-processing and summarization. Future work should explore long-context architectures and multi-step reasoning trajectories to better support the most intricate clinical and research workflows.

\section*{Acknowledgements}
We acknowledge funding support from the National Science Foundation (NSF) under grants IIS-2405974 and IIS-2339216, and from the National Institutes of Health (NIH) under grant R35GM157217.

\bibliography{custom}

\appendix

\section{Heuristic Filter Detail}\label{sec:appendix_heuristic}
In this section, we provide a more granular explanation of the heuristic filtering strategies employed during the API call synthesis and verification phase.

The specific filtering logic varies across the three integrated databases to account for differences in their API architectures and the nature of the biological data they provide. For UniProt, which primarily provides functional protein annotations and sequence data, we implement a strict deduplication process by filtering out all API calls targeting the same unique identifier, such as a UniRef entry ID or keyword entry ID, within the same tool to prevent the over-representation of specific proteins. Furthermore, we validate every execution result by discarding any responses that return empty lists or "null" search results, thereby ensuring that every retained API call contains at least one valid, non-empty biological observation. Ensembl requires a more nuanced dual-path approach to balance diversity and validity when handling complex genomic coordinates. For endpoints with a restricted set of valid parameter combinations (defined as fewer than 20), where strict ID deduplication would yield insufficient data, we selectively retain entries where the optional parameters, such as species or variants, are not identical, while query IDs are the same. Conversely, for "rich" APIs with an expansive range of possible inputs, we apply a strategy similar to UniProt by filtering out any samples where the combination of required parameters is identical to an existing entry to prevent the model from over-fitting to specific genomic regions. For NCBI, the strategy is optimized for high-throughput tools and general metadata retrieval. We apply specialized heuristics to the BLAST tool, only retaining parameter combinations that involve unique query sequences and return at least one significant alignment hit, while removing matchless queries that cannot support downstream scientific reasoning. Other NCBI APIs are filtered using a standard heuristic method that removes identical identifier calls and verifies that the retrieved observations remain biologically informative.

\section{Dataset Scale Analysis}\label{sec:appendix-datasize}

To examine how performance scales with training data volume, we train Qwen3-4B-Instruct~\cite{yang2025qwen3technicalreport} on six subsets of the \textsc{BioTool} training split, ranging from 10\% to 100\%. Table~\ref{tab:datasize} presents the improvement over the untuned Qwen3-4B-Instruct baseline, whose overall Exact Match, API Success, and BioTool Score are 3.6, 73.3, and 63.1, respectively. The results show that even the smallest training subset yields substantial gains across all three metrics, confirming that \textsc{BioTool} provides strong supervision from the early stages of scaling. As the training set expands, the gain in BioTool Score increases steadily, while Exact Match continues to improve throughout the full range, indicating that parameter-level precision remains the principal source of additional benefit at larger scales.

\begin{table}[t]
\centering
\setlength{\tabcolsep}{6pt}
\renewcommand{\arraystretch}{1.1}
\begin{tabular}{lccc}
\toprule
\textbf{Training Size} & \textbf{$\Delta$ EM} & \textbf{$\Delta$ AS} & \textbf{$\Delta$ BioTool} \\
\midrule
10\%  & 18.6 & 20.7 & 23.2 \\
20\%  & 27.3 & 21.3 & 25.5 \\
40\%  & 33.6 & 23.0 & 27.8 \\
60\%  & 36.2 & 23.1 & 28.5 \\
80\%  & 37.6 & 22.9 & 28.9 \\
100\% & \textbf{38.7} & \textbf{24.2} & \textbf{30.4} \\
\bottomrule
\end{tabular}
\caption{Overall performance gains of Qwen3-4B-Instruction fine-tuned on different fractions of the \textsc{BioTool} training set, measured as absolute improvements over the base Qwen3-4B-Instruction baseline. $\Delta$ EM, $\Delta$ AS, and $\Delta$ BioTool denote gains in Exact Match, API Success, and BioTool Score, respectively. Bold values indicate the largest gain in each column.}
\label{tab:datasize}
\end{table}

\section{Generalization Capability}\label{sec:appendix-api-split}

To examine whether \textsc{BioTool}-fine-tuned models can generalize beyond the APIs observed during training, we construct a stricter evaluation split based on API identity, such that all samples associated with the same API function are assigned to a single partition, and every API in the test set is unseen during training. We then compare the resulting performance of Qwen3-4B-Instruct against GPT-5.1 and GPT-5.1-Codex. As shown in Table~\ref{tab:api_split_results}, Qwen3-4B-Instruct retains a clear advantage in Exact Match and also achieves the strongest BioTool Score, indicating that \textsc{BioTool} fine-tuning continues to improve the structural fidelity and overall quality of API calling even when evaluation is conducted on previously unseen functions. Compared with the previously reported results under the standard random split, the performance gap becomes smaller in this setting, indicating that generalization to unseen APIs is substantially more challenging than generalization to new instances of previously observed APIs. Nevertheless, the continued advantage of Qwen3-4B-Instruct shows that \textsc{BioTool} fine-tuning yields transferable gains that extend beyond memorization of the training API set.

\begin{table}[t]
\centering
\setlength{\tabcolsep}{6pt}
\renewcommand{\arraystretch}{1.1}
\begin{tabular}{lccc}
\toprule
\textbf{Model} & \textbf{EM} & \textbf{AS} & \textbf{BioTool Score} \\
\midrule
GPT-5.1 & 4.2 & 88.6 & 76.5 \\
GPT-5.1-Codex & 3.8 & \textbf{94.7} & 83.5 \\
Qwen3-4B-Ins & \textbf{25.7} & 93.0 & \textbf{84.1} \\
\bottomrule
\end{tabular}
\caption{Model performance on the unseen-API evaluation split, where all instances associated with the same API function are assigned to a single partition. EM, AS, and BioTool Score denote Exact Match, API Success, and BioTool Score, respectively. Bold values indicate the best result in each column.}
\label{tab:api_split_results}
\end{table}

\section{Human Evaluation Details}
\label{sec:appendix-human-eval}

This section details the manual side-by-side assessment process and provides the raw preference data used to derive the winning rates reported in Section~\ref{sec:results_human_evaluation}. A total of 1,408 samples were evaluated for each comparison setting by researchers with biological backgrounds to compare the performance of tool-augmented models against the base GPT-5.1 generator. Table~\ref{tab:raw_human_eval} summarizes the distribution of these outcomes, including cases where both models performed well, or both failed to provide a satisfactory answer.

\begin{table}[ht]
\centering
\begin{tabular}{lrr}
\toprule
\textbf{Outcome} & \textbf{Qwen3-4B} & \textbf{Oracle} \\
\midrule
Total Samples & 1,408 & 1,408 \\
\midrule
Model A Wins & 75.3\% & 90.3\% \\
Model B Wins & 6.3\% & 1.8\% \\
Both Bad & 11.2\% & 3.3\% \\
Both Good & 7.2\% & 4.5\% \\
\bottomrule
\end{tabular}
\caption{Raw human preference distribution for pairwise model evaluations. Model A refers to the tool-augmented configuration (Qwen3-4B or Oracle), and Model B refers to the base GPT-5.1 model without tool access.}
\label{tab:raw_human_eval}
\end{table}

To provide a balanced comparison that accounts for samples where neither model showed a distinct advantage, we calculated the adjusted winning rate reported in Figure~\ref{fig:human_eval} based on the logic of McNemar's test~\cite{mcnemar1947note}. In this framework, ``Both Good'' and ``Both Bad'' responses are collectively treated as ties ($n_c = n_{good} + n_{bad}$) and adjusted by splitting them evenly between the two conditions. Specifically, given the raw preference counts $n_a$ and $n_b$, the adjusted preference numbers $n'_a$ and $n'_b$ were calculated as $n'_a = n_a + \frac{1}{2}n_c$ and $n'_b = n_b + \frac{1}{2}n_c$.

\paragraph{Annotation Reliability.}
To validate the quality of the labeling procedure, we computed inter-annotator agreement between the two human annotators on the manually labeled samples per task. Table~\ref{tab:iaa} reports Cohen's $\kappa$ and Krippendorff's $\alpha$ for each task and for the combined dataset. Both metrics exceed the commonly accepted threshold of 0.667 for ``acceptable'' agreement~\cite{krippendorff2011computing}, and fall within the ``substantial'' range ($\kappa \geq 0.61$) under the Landis \& Koch scale~\cite{landis1977measurement}, indicating reliable annotation across all settings.

\begin{table}[ht]
\centering
\begin{tabular}{lrrrr}
\toprule
\textbf{Model} & \textbf{N} & \textbf{\%Agree} & \textbf{$\kappa$} & \textbf{$\alpha$} \\
\midrule
Oracle    & 1,408 & 88.6\% & 0.800 & 0.800 \\
Qwen3-4B  & 1,408 & 82.0\% & 0.722 & 0.722 \\
\bottomrule
\end{tabular}
\caption{Inter-annotator agreement between two human annotators. Each comparison is between a tool-augmented model (Oracle or Qwen3-4B) and the base GPT-5.1 model without tool access. Agreement is computed on the four raw preference categories.}
\label{tab:iaa}
\end{table}

\newtcolorbox{promptcontainer}[1]{
    breakable,
    colback=gray!3,      
    colframe=gray!20,    
    coltitle=black,
    fonttitle=\bfseries\sffamily,
    title=#1,
    sharp corners,
    boxrule=0.5pt,
    left=3mm,
    right=3mm,
    top=3mm,
    bottom=3mm,
    middle=2mm,
    width=\columnwidth,
    fontupper=\rmfamily\small, 
    before skip=12pt,
    after skip=12pt
}

\section{Prompts}
\label{sec:appendix_prompt}

\subsection{Prompt for creating user queries}\label{sec:appendix_user_queries_prompt}

The following prompt is used to generate natural language user queries. It requires four distinct input streams: (1) the source document context, (2) API function specifications, (3) retrieved biological observations, and (4) in-context few-shot demonstrations.

\begin{promptcontainer}{System Prompt}
You generate realistic biomedical questions that researchers naturally ask.

\vspace{1em}
\noindent\textbf{TASK OVERVIEW: TWO-PHASE REASONING}

\medskip
\textbf{Phase 1 -- ANALYZE:} Map technical parameters to natural language concepts (Qualitative Mapping).

\textbf{Phase 2 -- GENERATE:} Create TWO questions (one Broad/Implicit, one Specific/Qualitative).

\vspace{1.5em}
\noindent\textbf{PHASE 1: PARAMETER MAPPING \& ABSTRACTION}

\medskip
Do not simply list parameters. You must translate \textit{Data} into \textit{Language}.

\begin{itemize}[leftmargin=*, itemsep=6pt, topsep=4pt]
    \item \textbf{VERBATIM (Keep Exact):} Unique identifiers (gene symbols, rsIDs, accessions), Raw sequences (FASTA format), and Coordinates (e.g., ``chr1:100-200'').
    \item \textbf{QUALITATIVE MAPPING (Translate Numbers/Codes):} 
        \begin{itemize}[label=$\circ$, noitemsep]
            \item \textbf{Thresholds:} Map high numbers to adjectives like ``strong'' or ``significant'' (e.g., \texttt{d\_prime=0.8} $\rightarrow$ ``strong linkage'').
            \item \textbf{Complex Codes:} Simplify technical strings to common terms (e.g., \texttt{1000GENOMES...} $\rightarrow$ ``1000 Genomes data'').
        \end{itemize}
    \item \textbf{IMPLICIT DEFAULTS (Selectively Omit):} If a parameter just ensures usable results (e.g., \texttt{format=json}), OMIT it. The user implies ``good results'' by asking.
\end{itemize}

\vspace{1.5em}
\noindent\textbf{PHASE 2: QUESTION GENERATION STRATEGY}

\medskip
Your goal is \textbf{Tool Bias}: The question should be specific enough that \textit{this tool} is the logical choice, without naming it.

\medskip
\textbf{Question 1: The ``Implicit'' Question (Natural \& Broad).} A question a biologist asks a colleague. Hide strict parameters; assume the tool's filters represent the broad intent.

\medskip
\textbf{Question 2: The ``Qualitative'' Question (Specific Demand).} A researcher asking for a specific \textit{quality}. Use adjectives to reflect parameter values (e.g., ``strong LD'').

\vspace{1em}
\textbf{RULES:}
\begin{itemize}[leftmargin=*, noitemsep]
    \item Ask only \textbf{one} question at a time.
    \item Keep the question concise and to the point.
    \item Do not use parentheses for supportive information.
\end{itemize}

\vspace{1em}
\noindent\textbf{OUTPUT FORMAT} \\
Return JSON only with keys: \texttt{param\_analysis}, \texttt{observation\_check}, and \texttt{questions}.
\end{promptcontainer}

\begin{promptcontainer}{User Prompt}
\textbf{GENERATE TWO NATURAL BIOMEDICAL QUESTIONS}

\vspace{1em}
\noindent\textbf{API DOCUMENTATION} \\
\textit{(Understand the tool's specific bias and domain:)} \\
\texttt{[API\_DOC\_TEXT]}

\vspace{1.5em}
\noindent\textbf{STEP 1 -- Classify parameters} \\
\textbf{PARAMS:} \\
\texttt{[PARAMS\_JSON]}

\vspace{1.5em}
\noindent\textbf{STEP 2 -- Check observation} \\
\textit{(Distinguish between AVAILABLE data and MISSING/EMPTY data)} \\
\textbf{OBSERVATION:} \\
\texttt{[OBSERVATION\_JSON]}

\vspace{1.5em}
\noindent\textbf{STEP 3 -- Write TWO questions}
\begin{itemize}[leftmargin=*, itemsep=4pt]
    \item \textbf{Question 1:} Broad intent (Natural tone, implies need for this specific tool).
    \item \textbf{Question 2:} Specific feature (Focus on a field that HAS data).
    \item \textbf{MUST include these identifiers:} \texttt{[IDENTIFIERS\_BLOCK]}
\end{itemize}

\vspace{1.5em}
\noindent\textbf{REFERENCE EXAMPLES} \\
\texttt{[FEW\_SHOTS\_TEXT]}

\vspace{1em}
Output JSON with \texttt{param\_analysis}, \texttt{observation\_check}, and \texttt{questions}.
\end{promptcontainer}

\subsection{Prompt for informative check}\label{sec:appendix_informative_check_prompt}

The following prompt is used to evaluate whether an observation is informative enough to answer a specific user query. It requires (1) the natural language user query and (2) the JSON representation of the biological observation.

\begin{promptcontainer}{System Prompt}
You are an evaluator for dataset filtering.

\vspace{1em}
\noindent\textbf{Goal:} Decide whether the OBSERVATION is informative (useful) for answering the USER QUERY.

\vspace{1em}
\noindent\textbf{IMPORTANT CONTEXT:} Observations are POSTPROCESSED SUMMARIES (often partial). This is NOT a strict completeness check.

\vspace{0.5em}
Be concise and deterministic.
\end{promptcontainer}

\begin{promptcontainer}{User Prompt}
\noindent\textbf{USER QUERY:} \\
\texttt{[USER\_QUERY\_TEXT]}

\vspace{1.5em}
\noindent\textbf{OBSERVATION (tool output / retrieved data):} \\
\texttt{[OBSERVATION]}

\vspace{1.5em}
\noindent\textbf{RUBRIC} \\

\begin{itemize}[leftmargin=*, itemsep=6pt, topsep=4pt]
    \item \textbf{informative=true} if the observation contains at least ONE relevant, non-trivial fact that can be used to answer part of the query without inventing details.
    \begin{itemize}[label=$\circ$, noitemsep]
        \item Examples/partial lists still count.
        \item Counts/aggregates/summaries still count.
        \item If the query asks ``which X'' but the observation only gives a count or a few examples, that is still \texttt{informative=true} (partial answer).
    \end{itemize}
    \item \textbf{informative=false} ONLY when:
    \begin{itemize}[label=$\circ$, noitemsep]
        \item Observation is an error / empty / placeholder, OR
        \item Observation content is clearly unrelated to the query intent, OR
        \item Observation is too vague to support even a single concrete statement relevant to the query.
    \end{itemize}
\end{itemize}

\vspace{1em}
\noindent\textbf{When writing the reason:}
\begin{itemize}[leftmargin=*, noitemsep]
    \item Focus on what CAN be answered using the observation (even partially).
    \item If partial, put the missing parts into limitations, but do NOT flip to false just because it's incomplete.
    \item Be concise and specific (name the fields/signals you used).
\end{itemize}

\vspace{1em}
\noindent\textbf{OUTPUT FORMAT} \\
\textit{(do NOT output JSON; output exactly these lines)} \\
INFORMATIVE: \texttt{true|false} \\
REASON: \texttt{<short reason>} \\
LIMITATIONS: \texttt{<optional; if none, write "none">}
\end{promptcontainer}

\subsection{Prompt for generating answers}
The following prompts are used to generate the final natural language responses for the human expert evaluation. These prompts require the original user query, the generated api call, and the corresponding biological observations as input. 

\begin{promptcontainer}{System Prompt (Base Model)}
You are a concise, accurate biomedical assistant. Answer the user question as directly as possible in 2--5 sentences, using your general biomedical knowledge and reasonable domain assumptions. Do NOT mention tools, APIs, databases, internet access, or that you cannot look things up. Do NOT tell the user how to get the information. Answer directly. If the question asks for record-level details you cannot know exactly, give the best plausible answer in a natural, helpful way without refusals or meta statements (avoid phrasing like ``I can't'', ``I don't know'', ``without risk of error'').
\end{promptcontainer}

\begin{promptcontainer}{User Prompt (Base Model)}
\texttt{[USER\_QUERY]}
\end{promptcontainer}

\begin{promptcontainer}{System Prompt (Tool-Augmented)}
You are a concise, accurate biomedical assistant. You are given a user question plus a tool call and its observation output. Answer in 2--6 sentences. Use the observation as primary evidence and your general knowledge. Write the answer directly as if you already know the facts. If the observation is insufficient, you may add general biomedical context, but do not invent record-level facts that should come from the observation.
\end{promptcontainer}

\begin{promptcontainer}{User Prompt (Tool-Augmented)}
\noindent\textbf{User question:} \\
\texttt{[USER\_QUERY]}

\vspace{1.5em}
\noindent\textbf{API call (for context):} \\
\texttt{[API\_CALL\_JSON]}

\vspace{1.5em}
\noindent\textbf{Observation (tool output):} \\
\texttt{[OBSERVATION\_TEXT]}
\end{promptcontainer}

\section{Tool and API List}\label{sec:appendix-tool}

\newtcolorbox{toolbox}[1]{
    breakable,
    colback=gray!3,
    colframe=gray!20,
    coltitle=black,
    fonttitle=\bfseries\sffamily,
    title=#1,
    sharp corners,
    boxrule=0.5pt,
    left=3mm,
    right=3mm,
    top=3mm,
    bottom=3mm,
    middle=2mm,
    width=\columnwidth,
    fontupper=\rmfamily\small,
    before skip=12pt,
    after skip=12pt
}

\setlist[itemize]{leftmargin=*, nosep}
\setlist[itemize,1]{label=$\bullet$}
\setlist[itemize,2]{label=$\circ$}

The following part enumerates all tools and their corresponding APIs used in this work, grouped by data source.

\begin{toolbox}{NCBI Tools}
\begin{itemize}
    \item \textbf{ESearch}
    \begin{itemize}
        \item esearch
    \end{itemize}
    
    \item \textbf{ELink}
    \begin{itemize}
        \item elink
    \end{itemize}
    
    \item \textbf{EFetch}
    \begin{itemize}
        \item efetch
    \end{itemize}
    
    \item \textbf{EInfo}
    \begin{itemize}
        \item einfo
    \end{itemize}
    
    \item \textbf{BLAST}
    \begin{itemize}
        \item blast
    \end{itemize}
\end{itemize}
\end{toolbox}

\begin{toolbox}{UniProt Tools}
\begin{itemize}
    \item \textbf{UniProtKB}
    \begin{itemize}
        \item get\_uniprotkb\_entry
        \item search\_uniprotkb
        \item stream\_uniprotkb
    \end{itemize}
    
    \item \textbf{UniRef}
    \begin{itemize}
        \item get\_uniref\_by\_id
        \item get\_uniref\_light
        \item get\_uniref\_members
        \item search\_uniref
        \item stream\_uniref
    \end{itemize}
    
    \item \textbf{UniParc}
    \begin{itemize}
        \item get\_uniparc\_by\_upi
        \item get\_uniparc\_databases
        \item get\_uniparc\_light
        \item search\_uniparc
        \item stream\_uniparc
    \end{itemize}
    
    \item \textbf{GeneCentric}
    \begin{itemize}
        \item get\_genecentric\_by\_accession
        \item get\_genecentric\_by\_proteome\_id
        \item search\_genecentric
        \item stream\_genecentric
    \end{itemize}
    
    \item \textbf{Proteomes}
    \begin{itemize}
        \item get\_proteome\_by\_upid
        \item search\_proteomes
        \item stream\_proteomes
    \end{itemize}
    
    \item \textbf{Literature citations}
    \begin{itemize}
        \item get\_citation\_by\_id
        \item search\_literature\_citations
        \item stream\_literature\_citations
    \end{itemize}
    
    \item \textbf{Keywords}
    \begin{itemize}
        \item get\_keyword\_by\_id
        \item search\_keywords
        \item stream\_keywords
    \end{itemize}
    
    \item \textbf{Human diseases}
    \begin{itemize}
        \item get\_disease\_by\_id
        \item search\_human\_diseases
        \item stream\_human\_diseases
    \end{itemize}
    
    \item \textbf{Subcellular locations}
    \begin{itemize}
        \item get\_location\_by\_id
        \item search\_subcellular\_locations
        \item stream\_subcellular\_locations
    \end{itemize}
    
    \item \textbf{Cross-referenced databases}
    \begin{itemize}
        \item get\_crossref\_database\_by\_id
        \item search\_crossref\_databases
        \item stream\_crossref\_databases
    \end{itemize}
    
    \item \textbf{Taxonomy}
    \begin{itemize}
        \item get\_taxonomy\_by\_id
        \item search\_taxonomy
        \item stream\_taxonomy
    \end{itemize}
    
    \item \textbf{UniRule}
    \begin{itemize}
        \item get\_unirule\_by\_id
        \item search\_unirule
        \item stream\_unirule
    \end{itemize}
    
    \item \textbf{ARBA}
    \begin{itemize}
        \item get\_arba\_by\_id
        \item search\_arba
        \item stream\_arba
    \end{itemize}
    
    \item \textbf{Archive}
    \begin{itemize}
        \item get\_archive\_id
    \end{itemize}
\end{itemize}
\end{toolbox}

\begin{toolbox}{Ensembl Tools}
\begin{itemize}
    \item \textbf{Comparative Genomics}
    \begin{itemize}
        \item get\_alignment\_region
        \item get\_cafe\_genetree\_by\_id
        \item get\_cafe\_genetree\_by\_member\_id
        \item get\_cafe\_genetree\_by\_member\_symbol
        \item get\_genetree\_by\_id
        \item get\_genetree\_member\_by\_id
        \item get\_genetree\_member\_by\_symbol
        \item get\_homology\_by\_id
        \item get\_homology\_by\_symbol
    \end{itemize}
    
    \item \textbf{Cross References}
    \begin{itemize}
        \item get\_xrefs\_by\_id
        \item get\_xrefs\_by\_symbol
        \item lookup\_xref\_name
    \end{itemize}
    
    \item \textbf{Information}
    \begin{itemize}
        \item get\_info\_analysis
        \item get\_info\_assembly
        \item get\_info\_assembly\_region
        \item get\_info\_biotypes
        \item get\_info\_biotypes\_groups
        \item get\_info\_biotypes\_name
        \item get\_info\_compara\_methods
        \item get\_info\_compara\_species\_sets
        \item get\_info\_external\_dbs
        \item get\_info\_genomes
        \item get\_info\_genomes\_accession
        \item get\_info\_genomes\_assembly
        \item get\_info\_genomes\_division
        \item get\_info\_genomes\_taxonomy
        \item get\_info\_species
        \item get\_info\_variation\_population\_name
        \item get\_info\_variation\_populations
        \item get\_info\_variation\_sources
    \end{itemize}
    
    \item \textbf{Linkage Disequilibrium}
    \begin{itemize}
        \item get\_ld\_around\_variant
        \item get\_ld\_pairwise
        \item get\_ld\_region
    \end{itemize}
    
    \item \textbf{Lookup}
    \begin{itemize}
        \item lookup\_by\_id
        \item lookup\_by\_symbol
    \end{itemize}
    
    \item \textbf{Mapping}
    \begin{itemize}
        \item map\_assembly
        \item map\_cdna\_to\_genome
        \item map\_cds\_to\_genome
        \item map\_translation\_to\_genome
    \end{itemize}
    
    \item \textbf{Ontologies and Taxonomy}
    \begin{itemize}
        \item get\_ontology\_ancestors
        \item get\_ontology\_ancestors\_chart
        \item get\_ontology\_descendants
        \item get\_ontology\_id
        \item get\_ontology\_name
        \item get\_taxonomy\_classification
        \item get\_taxonomy\_id
        \item get\_taxonomy\_name
    \end{itemize}
    
    \item \textbf{Overlap}
    \begin{itemize}
        \item overlap\_by\_id
        \item overlap\_by\_region
        \item overlap\_translation
    \end{itemize}
    
    \item \textbf{Phenotype annotations}
    \begin{itemize}
        \item get\_phenotype\_by\_accession
        \item get\_phenotype\_by\_gene
        \item get\_phenotype\_by\_region
        \item get\_phenotype\_by\_term
    \end{itemize}
    
    \item \textbf{Regulation}
    \begin{itemize}
        \item get\_binding\_matrix
    \end{itemize}
    
    \item \textbf{Sequence}
    \begin{itemize}
        \item get\_sequence\_by\_id
        \item get\_sequence\_by\_region
    \end{itemize}
    
    \item \textbf{Transcript Haplotypes}
    \begin{itemize}
        \item get\_transcript\_haplotypes
    \end{itemize}
    
    \item \textbf{VEP}
    \begin{itemize}
        \item vep\_by\_hgvs
        \item vep\_by\_id
        \item vep\_by\_region
    \end{itemize}
    
    \item \textbf{Variation}
    \begin{itemize}
        \item get\_variation
        \item get\_variation\_by\_pmcid
        \item get\_variation\_by\_pmid
        \item variant\_recoder
    \end{itemize}
    
    \item \textbf{Variation GA4GH}
    \begin{itemize}
        \item get\_ga4gh\_callsets
        \item get\_ga4gh\_datasets
        \item get\_ga4gh\_features
        \item get\_ga4gh\_featuresets
        \item get\_ga4gh\_references
        \item get\_ga4gh\_referencesets
        \item get\_ga4gh\_variantannotationsets
        \item get\_ga4gh\_variants
        \item get\_query\_beacon
    \end{itemize}
\end{itemize}
\end{toolbox}

\end{document}